\documentclass[conference]{IEEEtran}
\ifCLASSINFOpdf
  \usepackage[pdftex]{graphicx}
\else
\fi
\usepackage[cmex10]{amsmath}
\usepackage{subcaption}

\usepackage{rotating}

\usepackage{multirow}

\begin{document}
%
\title{Comparing Data-mining Algorithms Developed for Longitudinal Observational Databases}

\author{\IEEEauthorblockN{Jenna Reps, Jonathan M. Garibaldi, \\ Uwe Aickelin, Daniele Soria}
\IEEEauthorblockA{School of Computer Science, \\
University of Nottingham\\
Nottingham,  NG8 1BB\\
Email: \{jzr,jmg,uxa,dqs\} @cs.nott.ac.uk}
\and
\IEEEauthorblockN{Jack E. Gibson and Richard B. Hubbard}
\IEEEauthorblockA{Clinical Sciences Building\\
Nottingham City Hospital\\
Nottingham, NG5 1PB\\
Email: \{jack.gibson,richard.hubbard\} @nottingham.ac.uk }}


%


\maketitle

\begin{abstract}
Longitudinal observational databases have become a recent interest in the post marketing drug surveillance community due to their ability of presenting a new perspective for detecting negative side effects.  Algorithms mining longitudinal observation databases are not restricted by many of the limitations associated with the more conventional methods that have been developed for spontaneous reporting system databases.  In this paper we investigate the robustness of four recently developed algorithms that mine longitudinal observational databases by applying them to The Health Improvement Network (THIN) for six drugs with well document known negative side effects.  Our results show that none of the existing algorithms was able to consistently identify known adverse drug reactions above events related to the cause of the drug and no algorithm was superior.      
\end{abstract}


%
\IEEEpeerreviewmaketitle

\section{Introduction}
Medical drugs are prescribed frequently throughout the world but each time a patient takes a drug there is a risk of the patient developing a side effect, referred to as an adverse drug reaction (ADR).  The purpose of a prescription drug is to improve a patient's medical state, but ironically, sometimes ADRs can cause a patient's medical state to deteriorate.  To prevent this occurring it is important to know all the ADRs that can occur and to be able to identify patients that have a high risk for developing a specific ADR.  Obvious ADRs can often be found during clinical trials but the main purpose of a clinical trial is to determine the effectiveness of the drug being tested and not to identify all the possible ADRs.  Less obvious ADRs, long term usage ADRs, ADRs resulting from co-prescription of drugs or ADRs that occur in subgroups of the population that are underrepresented in clinical trials (for example children and pregnant females) can only be detected by continuously monitoring patients who are prescribed the drug after marketing, a process known as post marketing surveillance. 

The majority of the methods implemented for post marketing surveillance use a database known as the Spontaneous Reporting System (SRS) database containing voluntary reports of suspected drug/s and adverse drug event pairs \cite{Bate2009} \cite{Bate1998} \cite{DuMouchel1999} \cite{Puijenbroek2002}.  The SRS database is known to have duplicated, missing and incorrect entries \cite{Almenoff2005}.  It is also common for SRS databases to be prone to under-reporting as ADRs corresponding to less serious medical events may not be reported or ADRs that are very rare may never be suspected by anyone.  As a consequence of these issues, it may not be possible to identify all ADRs by mining the SRS databases or identification may only be possible after many thousands of patients have been prescribed the drug and had the ADR.  This is undesirable as many patients may die before a rare but fatal ADR is identified.  

A new type of medical database, known as the longitudinal observational database (LOD), has recently gained interest from the research community \cite{Wilke2011} for post marketing surveillance as it does not rely on voluntary reports and offers a new perspective for detecting ADRs.  One example of a LOD is The Health Improvement Network (THIN) database (www.epic-uk.org) that contains medical and prescription records for all registered patients at participating general practices in the UK.  General practitioners are required to enter all the medical events they are made aware of, so serious but rare ADRs are likely to be contained within the database.

There are currently four algorithms based on sequential pattern mining techniques that have been developed to mine LODs to detect ADRs but each algorithm has only been applied to one type of LOD and there has been no research to compare the four algorithms and identify conditions where one algorithm is preferable.  In this paper we applied the existing algorithms to the THIN database for six drugs with generally well known ADRs and compare the measure known as the Mean Average Precision (MAP) \cite{Zorych2011} that indicates how well each algorithm can rank known ADRs above medical events that are unlikely to be ADRs from a collection of events.  We also calculated the precision of each algorithm on the top 10 and 50 ranked events for each drug.      

The continuation of this paper is as follows.  Section two contains descriptions of the THIN database, the drugs investigated and the four existing algorithms, including information relating to the LODs each algorithm was previously applied to.  Details of the method use to compare the existing algorithms can be found in section three. Section four contains the results and is followed by a discussion of the implications of the results in section five.  The paper finishes with the conclusion in section six.

\section{Background}

\begin{table*}[t]
\centering
\caption{General information on the population of patients prescribed each drug that was investigated . 
Total is the number of prescriptions of the drug in the database and includes repeat prescriptions, First is the number of first time prescriptions of the drug and $13$ months is the number of prescriptions where the drug was recorded as being prescribed for the first time in $13$ months for a patient. The average age and gender ratio were calculated by considering all prescriptions of the drug.
}
\label{tab:pat}
\begin{tabular}{cccccc}  \\  
 Drug & Total & First  & 13 months &Average Age (St Dev) &Gender Ratio (F/M) \\ 
 \hline
Ciprofloxacin & $483\;217$& $277\;871$ & $322\;482$ & $57.25(19.95)$ & $1.28$\\
Norfloxacin & $30\;043$ &$15\;160$ &$17\;390$ & $59.23(19.74)$& $2.90$ \\
Doxepin & $73\;684$ & $7607$ & $8265$ & $64.46(16.22)$ &	$2.65$	\\
Nifedipine & $2\;905\;177$ &$144\;356$ &$154\;128$ & $69.70(12.04)$ & $1.09$ \\
Benzylpenicillin Sodium & $1217$ & $1003$ & $1048$ & $26.07(24.89)$ & $1.10$ \\
Glibenclamide & $418\;473$ & $15\;222$ & $16\;445$ & $67.83(11.38)$ & $0.82$ \\						
\end{tabular}
\end{table*}

\subsection{THIN Database}
 The THIN database consists of a collection of information obtained from participating UK general practices including information on patients such as their year of birth, gender and family connections and the demographics of the area they live in.  The database also contains temporal information detailing a patient's prescription and medical event histories since registration. For this comparison a database containing records from 495 general practices was used.  This subset of the THIN database contained approximately four million patients, over 358 million prescription entries and over 233 million medical event entries.

Each medical event is recorded in the database by a reference code known as a Read code. The Read codes used in the THIN database are an independent system designed specifically for primary care but every ICD-9-CM (International Classification of Diseases, Ninth Edition, Clinical Modification) code (or analogues) have a corresponding Read code \cite{Shephard2011}. The Read codes suffer from redundancy as different Read codes can correspond to the same medical event, for example there are $15$ Read codes for the medical event `vomiting' under a range of categories including `History/symptoms', `Symptoms, signs and ill-defined conditions', `Infections and parasitic diseases' and `Unspecified conditions'. 


A known issue of the THIN database is that it is common to have incorrect time stamps of medical events corresponding to newly register patients.  As patients can change general practices at any age, when they register they may have a history of events that a doctor needs to record. The term `registration event dropping' is used when historic or previously diagnosed events of newly registered patients are entered into the database.  For example, when a new patient first visits their doctor they may inform the doctor of a previously diagnosed chronic illness such as `diabetes'.  This medical event will then be input into the database with a date corresponding to the visit, rather than the actual date the patient was diagnosed with diabetes.  As the dates recorded for the `registration event drops' are frequently incorrect, including them in a research study will bias results. Research suggests that 'registration event dropping' is significantly reduced after a patient is registered for a year.  To prevent 'registration event dropping' biasing the results in this study, the first 12 months of medical history after registration are ignored for each patient as justified in \cite{Lewis2005}.

As patients can move to a different practice at any time (or die), in this study we only include prescriptions into the study where the corresponding patient is still active for a minimum of 30 days after.  This prevents the bias due to `under-reporting' of ADRs that may occur if a patient no longer attends the practice.  The last date a patient is active is considered to be the maximum date of any record for the patient or the patient's date of death.

\subsection{Drugs Investigated}
Six different drugs with variable attributes and prescription indications (cause of the prescription) were chosen for the investigation.  The drug Nifedipine is a calcium channel blocker that helps relax the smooth muscles in the heart and blood vessels allowing blood to flow with greater ease and is therefore used to treat prophylaxis of angina, hypertension or Raynaud�s phenomenon. The penicillin Benzylpenicillin Sodium and fluoroquinolones Ciprofloxacin and Norfloxacin are three antibiotics used to treat bacterial infections.  The other two drugs investigated are Doxepin a tricyclic antidepressant and Glibenclamide a sulfonylurea used to treat type 2 diabetes mellitus. 

The drugs Ciprofloxacin and Norfloxacin were chosen to investigate how good the data mining methods are when applied to different drug population sizes (the collection of patients prescribed the drug) as the majority of known ADRs are the same for both drugs but the number of prescriptions recorded in the database are $483\;217$ and $30\;043$ for Ciprofloxacin and Norfloxacin respectively.  The two drugs also have a similar average age of the patients when prescribed the drug but have different drug population gender distributions.  Nifedipine was previously chosen to investigate one of the existing algorithms, so we also chose to use Nifedipine in this study to gain some insight into how robust the existing algorithms are when applied to different LODs. 

Benzylpenicillin Sodium has the lowest number of prescriptions of any of the drugs chosen for the study, with only 1217 prescriptions being recorded in the database. Benzylpenicillin Sodium is also considered a fairly safe drug with only a few known ADRs being listed.  This will test how well the algorithms do with small amounts of data and fewer ADRs to detect.  It also has the lowest average age that the patients are prescribed the drug, 26 years, compared to the other drugs with average ages ranging from 57 years to 70 years.  Doxepin and Glibenclamide were chosen for variety as they are prescribed for different illnesses than the other drugs.       

Table \ref{tab:pat} presents the number of patients that are prescribed each drug in the database and lists general statistics such as the mean age and patient gender ratio (females/males) for each drug and algorithm pair.      

\subsection{Existing Algorithms}
\subsubsection{Algorithm 1}
Methods that are implemented on Spontaneous Reporting System (SRS) databases only have limited information as the drug prescription rates and background incident rates of medical events are both unknown.  To overcome these issues disproportionality methods are implemented as the calculations are independent of the medical event and drug prescription rates, due to these terms cancelling out during division.  The disproportionality methods make use of a contingency table, see Table \ref{tab:srs}.  For example, a disproportionality algorithm known as the Reporting Odds Ratio (ROR), first contrasts how often the event of interest occurs with the drug of interest compared to any other drug ($\frac{w_{00}}{w_{10}}$) and then compares this with the contrast of how often any other event occurs with the drug of interest compared to any other drug ($\frac{w_{01}}{w_{11}}$), see Eq. \ref{eq:ror}.

 \begin{equation}
\label{eq:ror}
ROR = \frac{w_{00}/w_{10}}{w_{01}/w_{11}}
\end{equation} 

Previous work has investigated applying algorithms developed for SRS databases after transforming a LOD into a SRS style database by inferring suspected drug and medical events pairs that are ADRs \cite{Zorych2011}.  Zorych {\it et al.} implemented three different ways to map the LOD into an SRS database, including a mapping called 'Modified-spontaneous reporting system' that incorporates the additional information on the number of patients that do not have any suspected ADRs after a drug and the background rate of medical events available in LODs.  They found that incorporating the addition information did not improve results in the simulated and real databases studied, consequently in this paper we chose to transform the THIN database using their 'SRS mapping' as this method is more efficient.  The values in the contingency table for drug $X$ and medical event $Y$ using the 'SRS mapping' are calculated such that:

\begin{description}
\item[$w_{00}$] is the number of distinct occurrences of event $Y$ within 30 days after the drug $X$ is prescribed
\item[$w_{01}$] is the number of distinct occurrences of non Y medical events that occur within 30 days after the drug $X$ is prescribed
\item[$w_{10}$] is the number of distinct occurrences of event $Y$ within 30 days after any drug other than $X$ is prescribed
\item[$w_{11}$] number of distinct occurrences of non $Y$ events within 30 days after any non $X$ drug is prescribed.  
\end{description}

The SRS disproportionality algorithm implemented in this paper is the ROR, where medical events are ranked in descending order of the left bound of the 90\% confidence interval of the ROR, Eq. \ref{eq:ror2}, as previous work showed that the $ROR_{05}$ was consistently better than the ROR \cite{Zorych2011}. 

\begin{equation}
\label{eq:ror2}
ROR_{05} = exp ( ln(\frac{w_{00}/w_{10}}{w_{01}/w_{11}}) \\ -1.645 \times \sqrt{ \frac{1}{w_{00}} +\frac{1}{w_{01}} +\frac{1}{w_{10}} +\frac{1}{w_{11}} })
\end{equation} 

\begin{table}
\centering
\caption{Contingency table used in existing SRS methods.}
\label{tab:srs}
\begin{tabular}{c|cc} 
          & Event j =Yes & Event j =No \\ \hline
Drug i=Yes & w$_{00}$ & w$_{01}$ \\
Drug i=No & w$_{10}$ & w$_{11}$ \\ 
\end{tabular}
\end{table} 

\subsubsection{Algorithm 2}
Noren {\it et al.} developed, specifically for LODs, a disproportionality based sequential pattern mining algorithm that uses temporal information contained in LODs to contrast the Observed to Expected ratio (OE ratio) of an event and drug pair between two different time periods \cite{Noren2010}.  The database the OE ratio was developed for is the UK IMS Disease Analyzer, a database containing UK general practice records containing over two million patients and 120 million prescriptions. The database contained $3 445$ drugs and $5 753$ medical events encoded by the ICD-10 \cite{ICD10}. When the UK IMS Disease Analyzer database was mined by the OE ratio for the drug Nifedipine it was found to have a $precision_{10}=0.7$.   

The OE ratio algorithm compares the number of patients that have the first prescription of drug $x$ (in thirteen months) followed by event $y$ within a set time $t$ relative to the expected number of patients if drug $x$ and event $y$ were independent.  Letting $n_{xy}^{t}$ denote the number of patients that have drug $x$ for the first time and event $y$ occurs within time period $t$, $n_{.y}^{t}$ denote the number of patients that are prescribed any drug for the first time and have event $y$ within time period $t$. $n_{x.}^{t}$ denote the number of patients that have drug $x$ for the first time with an active follow up in time period $t$ and $n_{..}^{t}$ denote the number of patients that have any drug for the first time with an active follow up in time period $t$.  The expected number of patients that have drug $x$ and then event $y$ in a time period $t$ is then,
\begin{equation}
E_{xy}^{t} = n_{x.}^{t} \frac{n_{.y}^{t}}{n_{..}^{t}}
\label{eq:expec}
\end{equation}    
If for a given drug, the event occurs more than expected, the ratio between the observed and expected will be greater than one.  By taking the $log_{2}$ of the ratio, a positive value suggests an interesting association between a drug and event. Modifying the equation to prevent the problem of rare events or drugs resulting in a small expectation that can cause volatility, a statistical shrinkage method is applied.  
\begin{equation}
IC = log_{2} \frac{n_{xy}^{t} + 1/2}{E_{xy}^{t} + 1/2}
\label{eq:icshrink}
\end{equation}
The shrinkage adds a bias for the $IC$ towards zero when an event or drug is rare. The credibility intervals for the $IC$ are the logarithm of the solution to Eq. \ref{eq:ci} with $q=0.025$ and $q=0.975$.
\begin{equation}
\int_{0}^{\mu_{q}} \frac{(E_{xy}^{t}+1/2)^{n_{xy}^{t}+1/2}}{\Gamma(n_{xy}^{t}+1/2)} u^{(n_{xy}^{t}+1/2)-1} e^{-(n_{xy}^{t}+1/2)} du = q
\label{eq:ci}
\end{equation}
The above can find possible drug and event associations of interest for a given $t$, however, the authors suggest that general temporal patterns can be found by comparing the $IC$ of two different time periods.  The follow-up period of primary interest is denoted by $u$ and the control time period by $v$.  This removes event and drug relationships that just happen to occur more in certain sub-populations.  The difference between the $IC$ for both time periods is,
\begin{equation}
log_{2} \frac{n_{xy}^{u}}{E_{xy}^{u}} - log_{2} \frac{n_{xy}^{v}}{E_{xy}^{v}}
\end{equation} 
re-arranging and adding a shrinkage term gives,
\begin{equation}
IC_{\Delta} = log_{2}  \frac{n_{xy}^{u}+1/2}{E_{xy}^{u*}+1/2}
\end{equation}
where
\begin{equation}
E_{xy}^{u*} = \frac{n_{xy}^{v}}{E_{xy}^{v}}.E_{xy}^{u} 
\end{equation}

In this paper we calculate the $IC_{\Delta}$ as described above by contrasting the 30 day period after the drug prescription with a time period of $27$ to $21$ months prior to prescription. The OE ratio ranks medical events in descending order of the $IC_{\Delta}$, but removes some noise by filtering medical events with a positive $IC$ a month prior to the prescription or with a positive $IC$ on the day of prescription.  As the THIN database does not contain information on the time during the day that a prescription is issued or a medical event is recorded, it is possible that medical events occurring on the same day as the prescription may be ADRs, so we investigate two different implementations of the OE ratio, the OE ratio 1 filters medical events with an $IC$ value a month prior to the prescription greater than the $IC$ value in the month after (not including the day of prescription) and OE ratio 2 filters every medical event with an $IC$ value a month prior to the prescription or on the day of prescription that is greater than the $IC$ value a month after.   

\subsubsection{Algorithm 3}
Mining Unexpected Temporary Association Rules given the Antecedent (MUTARA) \cite{Jin2006} is a sequential pattern mining algorithm that finds medical events that occur more than expected within a user defined time period after a drug is first prescribed.  MUTARA implements a measure of interestingness frequently used in sequential pattern mining known as Leverage.  In the context of the medical databases the Leverage gives an indication of how temporally dependent a medical event is on the the presence of a drug, as it is the number of patients that have the medical event after the first time they are prescribed the drug of interest minus the expected number of patients who would have the medical event if the presence of the medical event was independent of the presence of the drug.

MUTARA was developed to be originally implemented on the Queensland Linked Data Set (QLDS) comprising of the Commonwealth Medicare Benefits Scheme (MBS), Pharmaceutical Benefits Scheme (PBS) and Queensland Hospital morbidity data.  The database contained 2020 different diagnoses (medical events) and 758 distinct drug codes.  The database contains limited information on a patient's medical history, as it only contains information while a patient is in hospital.  When MUTARA was applied to the QLDS to detect ADR for older females prescribed alendronate the precision of the top ten events ($precision_{10}$) was $0.1$.
 
It is common for patients to have medical events repeated in their sequence and if a patient has a disease shortly before a prescription and then again within $T$ days of the prescription it is unlikely that the disease is an ADR.  As a consequence, the authors of MUTARA decided to filter `predictable' medical events by removing any medical events that occurred $T$ days after the drug prescription and also occurred in a user defined time period prior to the drug prescription. 

If we let $P(X \overset{T}{\hookrightarrow} Y)$ denote the probability of having event $Y$ `unpredictably' within $T$ days of drug $X$, then if event $Y$ occurs independent of drug $X$,  $P(X \overset{T}{\hookrightarrow} Y)=P(X).P( \overset{T}{\hookrightarrow} Y)$. A large value for $P(X \overset{T}{\hookrightarrow} Y)-P(X).P( \overset{T}{\hookrightarrow} Y)$ would then suggest a dependency of $Y$ on drug $X$, indicating $Y$ as a possible ADR.

The above measure can be estimated by, 

\begin{equation}
Unexlev = Supp(X \overset{T}{\hookrightarrow} Y)-\frac{Supp(X).Supp( \overset{T}{\hookrightarrow} Y)}{\mbox{Population}}
\end{equation}

Where,  

\begin{itemize}
\item $Supp(X \overset{T}{\hookrightarrow} Y)$ - the number of patients in the database that have the medical event $Y$ within $T$ days of the first time being prescribed drug $X$ and do not have medical event $Y$ in a user defined time period prior to $X$. 
\item $Supp(X)$ - the number of patients in the database that are prescribed the drug of interest. 
\item $Supp( \overset{T}{\hookrightarrow} Y)$ -  the number of patients who have never been prescribed drug $X$ and have medical event $Y$ in a randomly chosen time period of $T$ days plus $Supp(X \overset{T}{\hookrightarrow} Y)$. 
\item Population - the total number of patients
\end{itemize}

MUTARA calculates the Unexlev for each medical event input by the user (in this paper we input any medical event that occurs within 30 days of the first time the drug is prescribed for at least one patient) and returns a ranked list in descending order of the Unexlev. 

In this paper we use $T$=30 and chose to investigate two different time periods prior to the prescription that determine if a medical event is `predictable', 180 days and 60 days directly prior to the day the drug is first prescribed (MUTARA$_{180}$ and MUTARA$_{60}$ respectively).  


\subsubsection{Algorithm 4}

Highlighting UTARs Negating TARs (HUNT) is a modified version of MUTARA \cite{Jin2010}, originally developed and implemented on the QLDS with previous results of a $precision_{10}=0.3$ when applied to detect ADR for older females prescribed alendronate and a $precision_{10}=0.1$ when applied for older males.  MUTARA was found to have problems distinguishing between ADRs and therapeutic failures, as therapeutic failure medical events frequently occur after the drug is prescribed and have a high Unexlev value.  Both therapeutic failures and ADRs have a high Unexlev, but unlike ADRs, therapeutic failure medical events should also occur prior to the drug prescription for some patients.  The `predictable' filter should impact on therapeutic failure events but not ADRs, so the rank of therapeutic failure events can be reduced by comparing the Unexlev with the standard Leverage that calculates the temporal dependency of a medical event on a drug but does not filter `predictable' events, see Eq. \ref{eq:lev}. 

\begin{equation}
\label{eq:lev}
Leverage = Supp(X \overset{T}{\rightarrow} Y)-\frac{Supp(X).Supp( \overset{T}{\rightarrow} Y)}{\mbox{Population}}
\end{equation}

Where,  
\begin{itemize}
\item $Supp(X \overset{T}{\rightarrow} Y)$ - the number of patients in the database that have the medical event $Y$ within $T$ days of the first time being prescribed drug $X$. 
\item $Supp(X)$ - the number of patients in the database that are prescribed the drug of interest. 
\item $Supp( \overset{T}{\rightarrow} Y)$ -  the number of patients who have never been prescribed drug $X$ and have medical event $Y$ in a randomly chosen time period of $T$ days plus $Supp(X \overset{T}{\rightarrow} Y)$. 
\item Population - the total number of patients
\end{itemize}
HUNT calculates both the Unexlev and Leverage values, assigns each medical event two ranks ($RANK_{Unexlev}$ and $RANK_{Leverage}$) based on the Unexlev and Leverage values respectively in descending order and finally returns the list of medical events in decreasing order of the rank ratio ($RR$), 

\begin{equation}
RR = \frac{RANK_{Leverage}}{RANK_{Unexlev}}
\end{equation}

In this paper HUNT is implemented with the same parameters as described for MUTARA. 

\section{Comparison Method}

Each data mining algorithm is applied to the THIN database and a ranked list of medical events is returned.  The events are ranked in descending order of the association between the drug of interest and event, so events the algorithm has deemed more likely to be ADRs are ranked higher, an example of this can be seen in Table \ref{tab:list}.  The algorithm is then analysed by investigating how well it has ranked each of the known ADRs that have occurred in the returned ranked list.  The known ADRs are those that are listed in the British National Formulary (BNF) \cite{BNF} for the specific drug, or medical events of type `adverse reaction to drug x' or containing information about the continuation of the drug prescription. 

Given a ranked set of events, we calculate the Truth measure $y_{(i)}$ for the $i^{th}$ ranked event, by letting $y_{(i)}= 1$ if the event is a known ADR and $y_{(i)}=0$ otherwise, as shown in Table \ref{tab:list}.  Table \ref{tab:list} shows an example of a returned list containing five events ranked by an algorithm and the corresponding $y$ values.  Using the $y$ values we can then use the measures described below to compare the different algorithms. 

\begin{table}[t]
\centering
\caption{An example of the medical event list associated to a specific drug and ordered by one of the algorithms.}
\label{tab:list}

\begin{tabular}{ccccc}
Event &  Rank Score & ADR & $y_{(i)}$ &   \\ \hline
Event 1 & 2.34 & No & $y_{(1)}= 0$ & \\
Event 5 & 2.12 & Yes & $y_{(2)}= 1$ & precision$_{2}=1/2$ \\
Event 4 & 1.75 & Yes & $y_{(3)}= 1$ & precision$_{3}=2/3$ \\
Event 2 & 1.74 & No & $y_{(4)}= 0$ & \\
Event 3 & 0.68 & No & $y_{(5)}= 0$ & \\
\end{tabular}
\end{table}

\begin{table*}[t]
\centering
\caption{Precision$_{10}$ for the different drug and algorithm combinations.}
\label{tab:acc10}
\begin{tabular}{cccccccc}
\multirow{2}{*}{Drug} & \multicolumn{7}{c}{precision$_{10}$} \\
                      &{OE Ratio 1} & { OE Ratio 2} & { MUTARA$_{60}$} & { MUTARA$_{180}$} & { HUNT$_{60}$} & { HUNT$_{180}$} & {ROR$_{05}$} \\ \hline
Norfloxacin & 0 & 0 & 0.1 & 0.1 & 0 & 0 & 0 \\
Benzylpenicillin Sodium & 0 & 0 & 0 & 0 & 0.1 & 0 & 0 \\
Nifedipine & 0.3 & 0.4 & 0 & 0 & 0.4 & 0.5 & 0 \\
Doxepin & 0.1 & 0.1 & 0.2 & 0.4 & 0.2 & 0.3 & 0 \\
Glibenclamide & 0 & 0 & 0 & 0 & 0.1 & 0 & 0 \\
Ciprofloxacin & 0 & 0.1 & 0.2 & 0.2 & 0.2 & 0.1 & 0 \\ \hline
Mean (3dp) & 0.067 & 0.100 & 0.083  & 0.117  & 0.167  & 0.150  & 0 \\   
\end{tabular}
\end{table*}

\begin{table*}[t!]
\centering
\caption{Precision$_{50}$ for the different drug and algorithm combinations.}
\label{tab:acc50}
\begin{tabular}{cccccccc}
\multirow{2}{*}{Drug} & \multicolumn{7}{c}{precision$_{50}$} \\
                      &{OE Ratio 1} & { OE Ratio 2} & { MUTARA$_{60}$} & { MUTARA$_{180}$} & { HUNT$_{60}$} & { HUNT$_{180}$} & {ROR$_{05}$} \\ \hline
Norfloxacin & 0.02 & 0.04 & 0.1 & 0.1 & 0.08 & 0 & 0 \\
Benzylpenicillin Sodium & 0.08 & 0.06 & 0.08 & 0.08 & 0.04 & 0.02 & 0.04 \\
Nifedipine & 0.2 & 0.26 & 0.12 & 0.16 & 0.2 & 0.22 & 0 \\
Doxepin & 0.1 & 0.l & 0.12 & 0.14 & 0.18 & 0.22 & 0 \\
Glibenclamide & 0.02 & 0.02 & 0.02 & 0 & 0.04 & 0.02 & 0.06 \\
Ciprofloxacin & 0.02 & 0.02 & 0.12 & 0.14 & 0.10 & 0.08 & 0.02 \\ \hline
Mean (3dp) & 0.073 & 0.083 & 0.093 & 0.103 & 0.107 & 0.093 & 0.02  
\end{tabular}
\end{table*}


\subsection{Precision K}  

The precision$_{k}$ of each algorithm is defined as the fraction of known ADRs that occur in the top $k$ events of the list returned by each algorithm for a specific drug, see Eq. (\ref{acc}).
\begin{equation}
\mbox{Precision$_{k}$}= \frac{\sum_{i=1}^{k}y_{(i)} }{k}
\label{acc}
\end{equation}
In this study two $k$ values are investigated, $k=10$ and $k=50$. 
  
\subsection{Mean Average Precision}
The mean average precision (MAP) is a measure that can be used to determine how well an algorithm generally ranks the medical events associated to a drug. This measure has previously been applied to compare SRS data mining algorithms adapted to be implemented on a LOD \cite{Zorych2011} and was also used in the Observational Medical Outcomes Partnership (OMOP) Cup, a competition where contestants were required to detect ADRs in a generated LOD \cite{OMOP}. 

The MAP is calculated by finding the average precision$_{k}$ for each $k$ corresponding to a known ADR,   
\begin{equation}
\mbox{MAP}= \frac{\sum_{K:y_{(K)}=1} \mbox{precision}_{K}}{\sum_{i} y_{(i)}}
\label{map}
\end{equation}

Using Table \ref{tab:list} as an example, as there are two known ADRs returned ($\sum_{i} y_{(i)}=2$) and the known ADRs in the table are ranked second and third we have $\{K:y_{(K)}=1\}=\{2,3\}$, so the MAP score is, 

\begin{equation}
\mbox{MAP}=\frac{\mbox{precision}_{2}+\mbox{precision}_{3}}{2}= \frac{1/2+2/3}{2}= \frac{7}{12}
\end{equation}

It was also possible to investigate how well each algorithm ranks the known ADRs depending on how common they are.  As the BNF states the risk of each known ADR by separately listing frequently, less frequently and rarely occurring known ADRs we also calculated the MAP score of each algorithm when only considering rarely occurring known ADRs (as unknown ADRs are likely to be rare).  We calculate the MAP scores for three different situations; considering all known ADRs, only considering rare known ADRs and lastly, only considering Read codes mentioning `adverse reaction to drug x' or a change in prescription.  

\section{Results}

\subsection{Precision$_{k}$}     
It can be seen in Table \ref{tab:acc10} that overall HUNT$_{60}$ had the highest precision$_{10}$ and the worse performing is the ROR$_{05}$.  At a 1\% significance level there was not enough evidence to show any of the algorithms had a significantly greater precision$_{10}$ than any other algorithm (one sided paired Mann-Witney U-test with multiple testing correction, p=0.01). 

HUNT$_{60}$ also had the greatest average precision$_{50}$, see Table \ref{tab:acc50}. Similarly to the precision$_{10}$, none of the algorithms had a significantly greater precision$_{50}$ (one sided paired Mann-Whitney U-test with multiple testing correction, p=0.01).  

\subsection{Mean Average Precision} 

\begin{figure*}[htp]
  \centering
  \begin{minipage}[t]{\textwidth}
    \centering
    \begin{minipage}[t]{\textwidth}
      \centering \includegraphics[width=\textwidth]{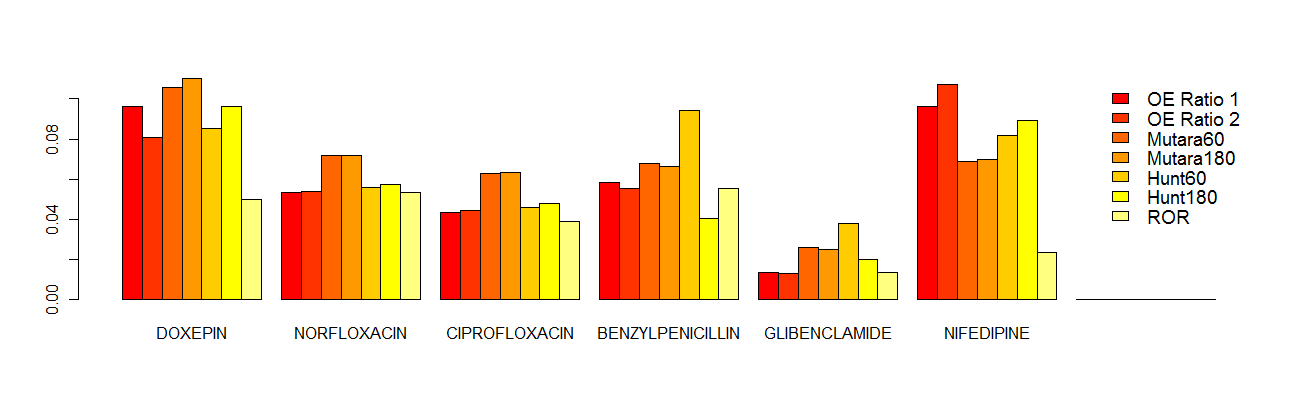} \subcaption{The general MAP scores for the different algorithms applied to the range of drugs.} \label{res:mapgen}
    \end{minipage}
    \begin{minipage}[b]{\textwidth}
      \centering \includegraphics[width=\textwidth]{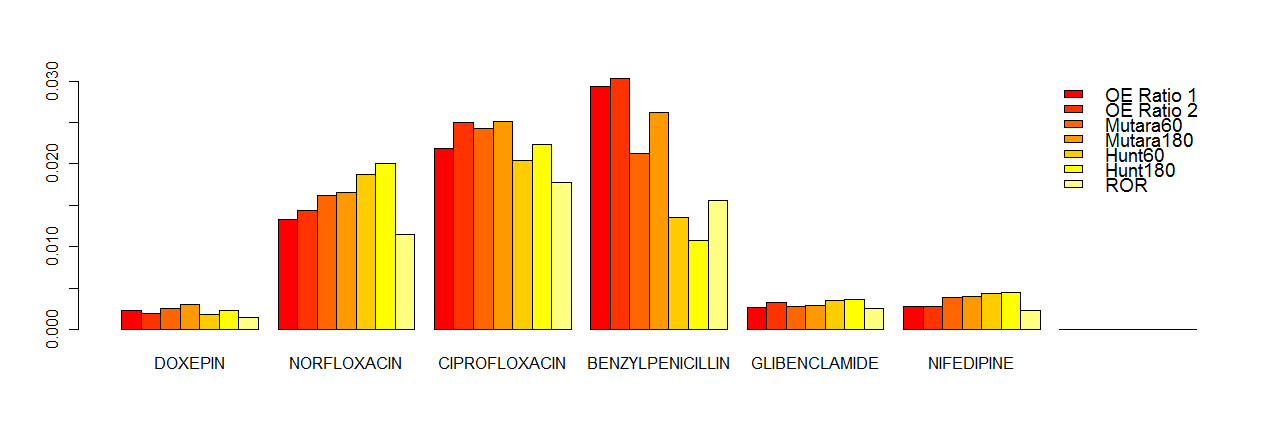} \subcaption{The MAP scores corresponding to the rare known ADEs for the different algorithms applied to the range of drugs.} \label{res:maprare}
    \end{minipage}
    \begin{minipage}[b]{\textwidth}
      \centering \includegraphics[width=\textwidth]{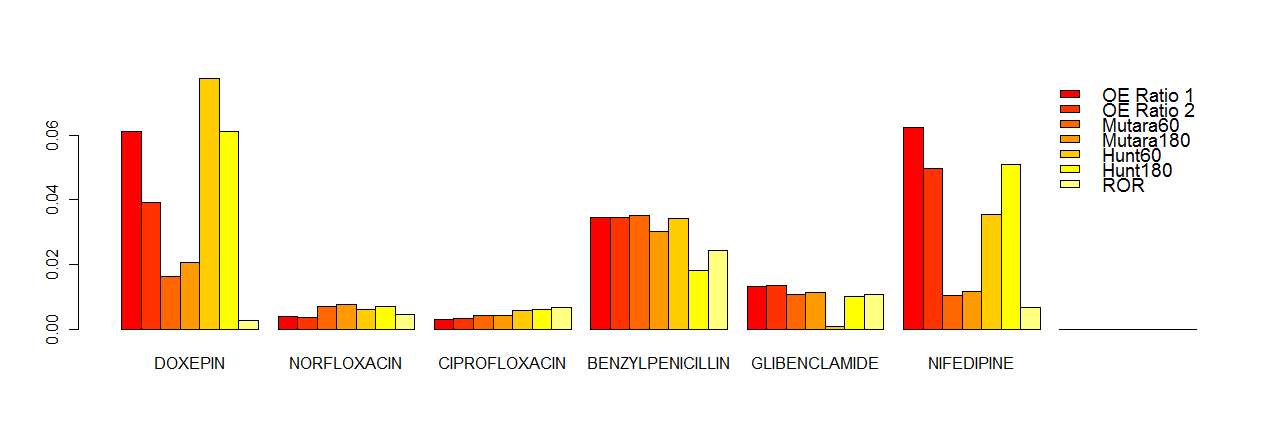} \subcaption{The MAP scores rating the algorithms ability to rank Read codes specifying `adverse drug reaction to x' or indicating a change in prescription applied to the range of drugs.} \label{res:mapdrug}
    \end{minipage}
  \end{minipage}

  \caption{Bar charts of the algorithm MAP scores.}
  \label{ig:animals}
\end{figure*}

Fig. \ref{res:mapgen} displays the general MAP scores using all the known ADEs listed on the BNF website for the different algorithms applied to the different drugs.  The results can be split into three different drug groups.  The first group of drugs, Doxepin, Norfloxacin and Ciprofloxacin all show a similar trend with MUTARA (MUTARA$_{60}$ and MUTARA$_{180}$) performing the best. The next group, Glibenclamide and Benzylpenicillin Sodium, show MUTARA outperforming the OE ratio but with HUNT$_{60}$ being the best performing algorithm. Lastly, the remaining group consisting of just one drug, Nifedipine, shows a unique trend with MUTARA performing relatively poorly and the OE ratio algorithm producing the best result.  The MAP scores for all the drugs apart from Glibenclamide range from $0.04-0.18$, with the MAP scores of Glibenclamide ranging from $0.01-0.04$.

Fig. \ref{res:maprare} shows the MAP scores for the algorithms when only considering rare known ADEs. The MAP scores for all the algorithms range between $0-0.03$.  The group of drugs Doxepin, Glibenclamide and Nifedipine have MAP scores less than $0.005$ whereas Norfloxacin, Ciprofloxacin and Benzylpenicillin Sodium have MAP scores between $0.01-0.03$. There was no algorithm that had a consistently greater MAP score over all the drugs, but the $ROR_{05}$ performed the worse for five out of the six drugs investigated. 

\section{Discussion}

The results show that all the existing algorithms have difficulty detecting rare known ADRs as, for each algorithm, the MAP score when only considering rare ADRs was much lower than the corresponding MAP score when including all the known ADRs.  It can also be observed that the MAP scores for Glibenclamide a drug from the class of medications called sulfonylureas that are known for rarely causing ADRs was lower than for all the other drugs.  The ability of the existing algorithms in detecting rare ADRs needs to be improved for them to be able to offer new information that cannot be found by mining SRS databases.  Another limitation with the existing algorithms is the variety of parameters that may need to change depending on the drug being studied and an improvement for each algorithm would be to develop a way to learn the suitable parameters dependent of the type of drug being investigated.

Fig \ref{res:mapgen} shows two important results, firstly that the algorithms are not robust, as their MAP scores varied across the different drugs and secondly that no algorithm was consistently better than the rest.  As a consequence, at current, it would be better to apply all the algorithms when mining LODs to detect ADRs and further investigate possible ADRs that are highly ranked by the majority of the algorithms.

Both MUTARA and HUNT estimate the background rate of a medical event by calculating how often the medical event occurs during some random time period for each patient that was not prescribed the drug.  As the time period chosen for each non drug patient will change every time the algorithm is implemented, the results of MUTARA and HUNT will be different.  In this study we did not investigate how much the MAP scores vary because of this and this is something that needs further investigation.  MUTARA and HUNT were developed to be applied to a patient population of a specified age group and gender but in this study we applied them to any age and gender as partitioning the drug population often causes an increase in the time it takes before rare ADRs are detected.  By applying MUTARA and HUNT to the whole drug population we increased the effects of confounding as the non drug population are likely to have different age and gender distributions.  Interestingly, MUTARA and HUNT performed better for Nifedipine than Norfloxacin and Ciprofloxacin even though there was less gender bias in the population of patients prescribed Nifedipine, this suggests that MUTARA and HUNT may still be suitable even when the drug patient population are a wide mix of different ages and genders. A final observation of the HUNT and MUTARA algorithms is the effect that the length of time prior to prescription used to filter `predictable' medical events has on the ability of HUNT and MUTARA to identify ADRs that have a high background rate.  For the drugs Doxepin, Norfloxacin, Ciprofloxacin and Nifedipine filtering medical events that occurred 180 days prior to the prescription improved the algorithms ability to detect ADRs but this was not the case for Benzylpenicillin sodium and Glibenclamide, where HUNT$_{60}$ perform much better than HUNT$_{180}$.  A likely reason for this is that many of the known ADRs for Benzylpenicillin sodium and Glibenclamide are generally common medical events and the longer the time period investigated prior to the prescription the more likely the common ADRs are to be filtered, this suggests that the algorithms should be implemented with multiple time period used for filtering as we do not know if the medical events that are unknown ADRs have a high prevalence or not.  

The case series approach of the OE ratio, comparing the same population at two different periods of time, reduces the confounding of age, gender and medical state but still has issues as a patient's medical state may have drastically changed over the two years and medical events that are related to the cause of the prescription are only likely to occur after the drug and not prior to the prescription.  If event relationships could be learned, the filters implemented by the OE ratio could not only filter medical events that occur more than expected on the day of prescription or a month prior to prescription but could also filter temporally related medical events.  We also stated that we wanted to investigate the possible issue of filtering medical events that occur more than expected on the day of prescription, as ADRs may be reported the same day and be incorrectly filtered.  By looking at Fig \ref{res:mapdrug} we can see that the OE ratio 1 generally performed better than the OE ratio 2 for medical events corresponding to `had adverse reaction' or `patient's prescription changed' suggesting filtering medical events that occur more than expected on the day of prescription may prevent detection of some ADRs.

The algorithms appear to be robust over different LODs, as the precision$_{10}$ results of the MUTARA$_{180}$/HUNT$_{180}$ applied to THIN averaged $0.117$ and $0.150$ respectively and were previously found to be $0.1$ and $0.3$/$0.1$ respectively.  The OE ratio applied to Nifedipine had a lower precision$_{10}$$=0.4$ when applied to THIN, however the authors of the OE ratio stated that when they implemented the algorithm they removed administrative events resulting in a precision$_{10}=0.7$ for Nifedipine, and although we ignored Read codes under the administration category, other Read codes that were included may be considered administration events, ignoring any event that could be considered an administration event resulted in a more comparable precision$_{10}=0.6$ when the OE ratio 2 was applied to THIN.

\section{Conclusion}
We applied four existing algorithms developed to detect ADRs by mining LODs by implementing them on the THIN database.  Our results show that no algorithm outcompeted the others, the existing methods had difficulty detecting rare ADRs and none of the algorithms was robust over all the drugs investigated.  Future work needs to investigate ways to learn suitable parameters for each algorithm depending on the drug being investigated and ways to filter illness progression and therapeutic failure events.





\bibliographystyle{IEEEtran}
\bibliography{bib/LitRevRef2_3authors,bib/LitRevRef_extras}
%



\end{document}